\documentclass{article}

\usepackage{amsmath,amssymb,amsfonts}
\usepackage{fontawesome5}
\usepackage{algorithm}
\usepackage{algorithmic}
\usepackage{graphicx}
\usepackage{textcomp}
\usepackage{multirow} 
\usepackage{makecell} 
\usepackage{booktabs}
\usepackage{threeparttable}
\usepackage{bbm}
\usepackage{mathtools}
\usepackage{bbold}
\usepackage{caption}
\usepackage{marvosym}
\usepackage{ifsym}
\usepackage[dvipsnames]{xcolor}        % colors
\newcommand{\ie}{\textit{i.e.}, }
\newcommand{\eg}{\textit{e.g.}, }

\newcommand{\wrt}{\textit{w.r.t. }}
\usepackage[colorlinks,
            linkcolor=red,
            anchorcolor=red,
            citecolor=blue,
            ]{hyperref}     % hyperlinks

\usepackage[T1]{fontenc}

\usepackage[preprint]{neurips_2024}

\title{Quality Sentinel: Estimating Label Quality and Errors in Medical Segmentation Datasets}

\author{%
  Yixiong Chen, ~Zongwei Zhou\thanks{Corresponding author: Zongwei Zhou (zzhou82@jh.edu)}, ~Alan Yuille \\
  Johns Hopkins University\\
}

\begin{document}

\maketitle

\begin{abstract}
  An increasing number of public datasets have shown a transformative impact on automated medical segmentation. However, these datasets are often with varying label quality, ranging from manual expert annotations to AI-generated pseudo-annotations. There is no systematic, reliable, and automatic quality control (QC). To fill in this bridge, we introduce a regression model, Quality Sentinel, to estimate label quality compared with manual annotations in medical segmentation datasets. This regression model was trained on over 4 million image-label pairs created by us. Each pair presents a varying but quantified label quality based on manual annotations, which enable us to predict the label quality of any image-label pairs in the inference. Our Quality Sentinel can predict the label quality of 142 body structures. The predicted label quality quantified by Dice Similarity Coefficient (DSC) shares a strong correlation with ground truth quality, with a positive correlation coefficient ($r=0.902$). Quality Sentinel has found multiple impactful use cases. (I) We evaluated label quality in publicly available datasets, where quality highly varies across different datasets. Our analysis also uncovers that male and younger subjects exhibit significantly higher quality. (II) We identified and corrected poorly annotated labels, achieving 1/3 reduction in annotation costs with optimal budgeting on TotalSegmentator. (III) We enhanced AI training efficiency and performance by focusing on high-quality pseudo labels, resulting in a 33\%--88\% performance boost over entropy-based methods, with a cost of 31\% time and 4.5\% memory. 
  The \href{https://github.com/Schuture/Quality-Sentinel/tree/main}{data and model} are released.
\end{abstract}

\section{Introduction}

Recent advances in large-scale organ segmentation \citep{liu2023clip} have shown significant potential for developing foundational models for general medical segmentation tasks. The datasets \citep{wasserthal2023totalsegmentator,jaus2023towards,qu2024abdomenatlas}, which comprise thousands of 3D scans covering numerous organs, facilitate the training of powerful and generalizable models. However, the annotation process for most of these datasets involves AI-assisted methodologies \citep{budd2021survey}, resulting in a predominance of model-generated pseudo labels. The inherent label noise can adversely affect the scalability and performance of the datasets. The models trained on these datasets may also lack reliability. Consequently, there is a critical need for a quality control (QC) method to make the annotation quality accessible, so that both the data quality and model performance can be improved.

The primary approach to controlling label quality involves radiologists' manual inspection. However, this is subjective and becomes increasingly time-consuming and impractical with the data growth, such as the AbdomenAtlas 1.1 \citep{li2023well}, which contains 3.7 million CT slices. Automatic methods for estimating pseudo label quality typically rely on the entropy of model predictions \citep{culotta2005reducing,mahapatra2018efficient} (\ie uncertainty) and the inconsistency of prediction among multiple models \citep{li2013adaptive,gal2017deep} (\ie diversity). However, these methods present significant drawbacks. Firstly, a model's feedback reflects its own confidence rather than the actual label quality. Secondly, the metrics are class-dependent (\eg organs with larger volumes exhibit higher entropy), which complicates the comparison of label quality across classes.

To address the challenges of accurately and efficiently measuring label quality under a uniform standard across all classes, we have devised a framework that trains models specifically for estimating the quality of labels (masks) in organ segmentation tasks. This framework employs the DSC, a comparison between the current label and the gold standard (ground truth), as the predictive metric for label quality. We refer to the models trained within this framework as \textbf{Quality Sentinel}. In previous literature on label evaluation models \citep{huang2016qualitynet,devries2018leveraging,galdran2018no,zhang2021quality,zhou2023volumetric}, the common practice is using an image-label pair as the input and predicting a quality metric. They are typically trained on a single or small collection of organs, hindering their practical use on modern multi-organ segmentation datasets. What sets Quality Sentinel apart from existing methods is it incorporates large language models to identify target organs, enhancing its performance in images featuring multiple organs. Specifically, we utilize the pretrained Contrastive Language-Image Pre-training (CLIP) \citep{radford2021learning} text encoder to embed the text description of 142 organs. This serves as a conditional input, allowing the model to recognize the organ it is evaluating. Furthermore, we introduce a scale-invariant ranking loss to overcome the challenges associated with training Quality Sentinel, a regressor, on a skewed DSC distribution. This not only facilitates smoother training but also significantly boosts its ranking accuracy.
In summary, our work presents three major \textbf{contributions} to the field of medical image segmentation and dataset evaluation. 

\begin{itemize}
    \item First, we introduce Quality Sentinel, a novel medical segmentation evaluation model designed to estimate label quality through DSC prediction. This model is distinguished by its use of text embeddings as conditions for different classes and a novel ranking loss mechanism, enabling effective multi-organ segmentation evaluation (Sect. \ref{sect:method}, Fig. \ref{fig:Quality Sentinel}).

    \item Second, we apply Quality Sentinel to reveal a large label quality variability among datasets. It also uncovers notable gender and age disparities in annotation quality. Specifically, labels associated with female and older patients exhibit lower quality, highlighting the need for increased diligence in dataset construction to mitigate the annotation biases, especially for these groups (Sect. \ref{sect:data_eval}, Tab. \ref{tab:eval}, Fig. \ref{fig:bias}).

    \item Third, we demonstrate the efficacy of Quality Sentinel as a sample selector for training segmentation models. It outperforms traditional diversity and uncertainty-based methods. On the one hand, it enhances the data efficiency in the human-in-the-loop process (active learning) for dataset development. On the other hand, by selecting high-quality pseudo labels in semi-supervised training, the segmentation models can be more reliable and powerful (Sect. \ref{sect:efficient_training}, Fig. \ref{fig:active}, Tab. \ref{tab:semi}).
\end{itemize}

\section{Related Work}

\subsection{Multi-Organ Segmentation}
With the rapid advances of deep segmentation models \citep{ronneberger2015u,isensee2021nnu,oktay2018attention,chen2021transunet,hatamizadeh2022unetr,tang2022self} and the emergence of large-scale multi-organ segmentation datasets \citep{ji2022amos,ma2021abdomenct,wasserthal2023totalsegmentator}, the need for data processing, filtering, and corresponding model training strategies has also grown. The existing dataset with the largest sample size is AbdomenAtlas \citep{qu2024abdomenatlas}. It labels 32 organs and tumors over 9,000 3D CT volumes with a human-in-the-loop technique. This method accelerates the annotation process to a large extent (as reported, the dataset was annotated within three weeks), but the high ratio of AI-generated pseudo labels may also make the dataset potentially risky in low-quality annotations. TotalSegmentator \citep{wasserthal2023totalsegmentator}, a large dataset with 1,000+ samples and 100+ classes, relies upon quality insurance via 3D renderings and expert checks. The labor-intensive checking results in the final checking rate of only 100 CT volumes. To this end, an efficient QC tool to ensure annotation quality is in urgent need.

\subsection{Medical Segmentation Quality Control}
QC tools raise a flag when the segmentation model under analysis incurs a lack of reliability or robustness \citep{galati2022accuracy}. They can be categorized into two folds: the pre-analysis and the post-analysis QC.  Pre-analysis QC \citep{lorch2017automated,tarroni2018learning,oksuz2019automatic} performs on the input of the segmentation model, it emulates expert criteria and detects low-quality samples (\eg incomplete or corrupted scans). By giving binary feedback, it can discard those samples to improve the robustness of the segmentation pipeline. Post-analysis QC \citep{alba2018automatic,robinson2018real,lin2022novel} focuses on the assessment of the segmentation outputs of a model. It usually infers common segmentation metrics (\eg DSC, Hausdorff Distance (HD), or uncertainty estimates), thus detecting a malfunction of models. Quality Sentinel is a tool that can serve as both pre-analysis and post-analysis. On the one hand, it helps to detect flaws in the existing datasets by analysis of the label quality. On the other hand, it can evaluate model outputs and find the weaknesses of the model, which can be beneficial for further refinement. Moreover, existing QC techniques primarily work on single organs (\eg Cardiac MR \citep{lorch2017automated,tarroni2018learning,oksuz2019automatic,alba2018automatic,robinson2018real}) or a small set of relevant organs (\eg knee and calf muscle \citep{zaman2023segmentation}). Quality Sentinel is an exploration of general-purpose QC tool that works for a wide variety of organs simultaneously.

\subsection{Language-driven Medical Image Analysis}
Large language models \citep{devlin2019bert,brown2020language} widely succeed in language processing and understanding. Visual language models (VLMs) \citep{wang2022medclip} such as CLIP have been explored to promote medical image analysis tasks such as medical vision question answering \citep{eslami2023pubmedclip}, medical object recognition \citep{qin2022medical}, and organ/tumor segmentation \citep{liu2023clip,ulrich2023multitalent}. VLM could be useful in these scenarios because the language encoder can generate intrinsic semantics of the anatomical structures for tasks in the medical domain with carefully designed medical prompts. Therefore, the vision module can be augmented with knowledge from the language domain. For Quality Sentinel, the text embedding acts as the condition for the vision representation of a target class, and similar classes share similar knowledge from language.

\section{Quality Sentinel}
\label{sect:method}

\begin{figure}[t]
\vspace{0cm}                          
\centering\centerline{\includegraphics[width=1.0\linewidth]{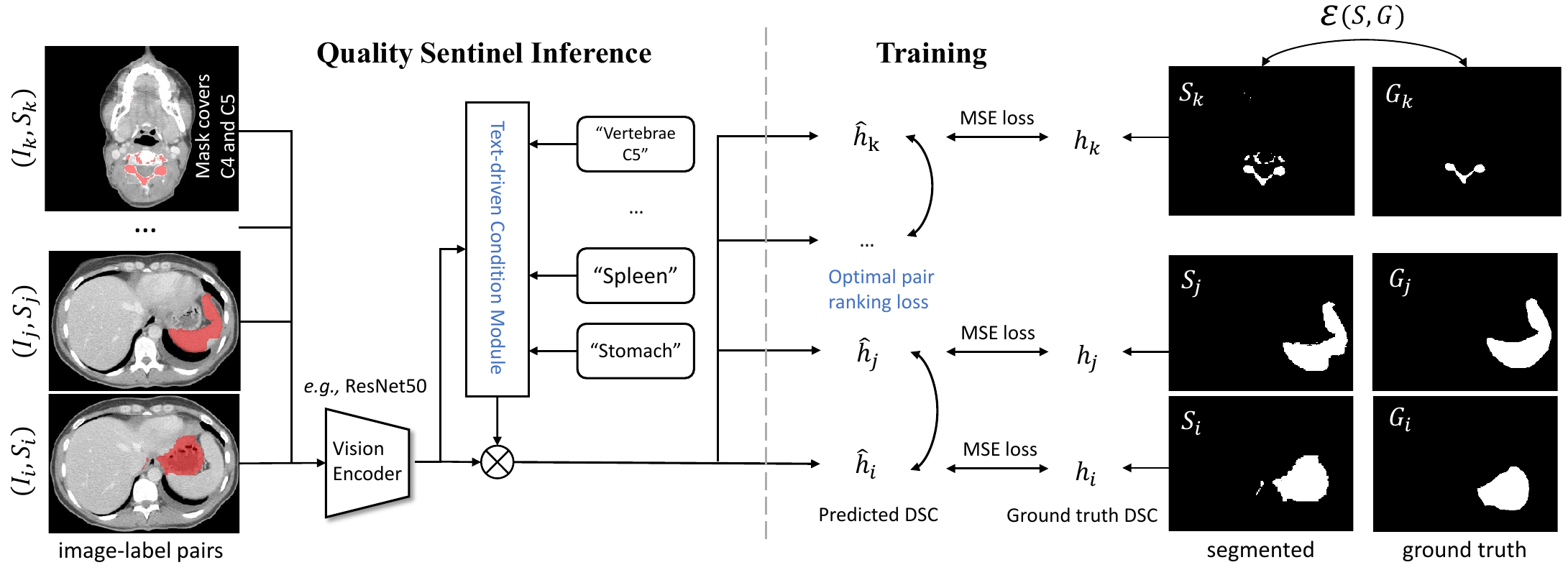}}
\caption{Overview of Quality Sentinel: Inputs are image-label pairs with pseudo labels. It employs a vision encoder to estimate DSC relative to ground truths. The Text-driven Condition Module encodes texts to augment the model for organ identification, addressing the ambiguity of organ-label matching (\eg for $(I_k, S_k)$ at upper left, the mask covers two organs). Training involves a compositional loss, combining optimal pair ranking and MSE, to align predicted with actual DSC.}
\label{fig:Quality Sentinel}
\end{figure}

\subsection{Formulation of the Quality Estimation Framework}
The Quality Sentinel framework, depicted in Fig. \ref{fig:Quality Sentinel}, evaluates data batches of $N$ samples, denoted as  $\mathcal{D}  =\{I_i, S_i, G_i\}_{i=1}^N$, where $I$, $S$, and $G$ represent an image, a segmented mask (pseudo label), and a ground truth label, respectively. The estimation model $\mathcal{Q}$, parameterized by $\theta$, estimates the DSC $\hat{h}_i = \mathcal{Q}(I_i, S_i; \theta)$, compared against the actual DSC $h_i = \mathcal{E}(S_i, G_i)$. To align the predicted $\hat{h}$ with the actual $h$, we introduce a compositional loss function:
\begin{equation}
    \mathcal{L}=\frac{1}{N}\sum_{i=1}^N\mathcal{L}_{MSE}(\hat{h}_i, h_i)+\lambda \mathcal{L}_{Rank}(\hat{\textbf{h}}, \textbf{h}),
\label{eq1}
\end{equation}
with $\lambda$ as a scaling constant. The mean square error (MSE) loss $\mathcal{L}_{MSE}(\hat{h}_i, h_i)$ aims to match the model's output with the ground truth DSC. The novel optimal-pair ranking loss $\mathcal{L}_{Rank}(\hat{\textbf{h}}, \textbf{h})$ enhances the model's ranking capability by optimizing the order of predicted and actual DSC within the batch $\mathcal{D}$.

For practical application, we process a 2D image-label pair as input to $\mathcal{Q}$, averaging estimates from 10 uniformly sampled 2D slices along the z-axis for 3D organ label quality estimation. This 2D-based method is superior to 3D methods because slice-level prediction is more accurate and can provide clinicians with specific information related to the single slices \citep{fournel2021medical}.

\subsection{Text-driven Condition Module}
Challenges arise with expanded or fragmented masks during the segmentation of small or irregularly shaped organs, as seen in the upper left of Fig. \ref{fig:Quality Sentinel} (a mask covers two organs), causing ambiguity in organ-label matching. This ambiguity compromised the evaluation model's accuracy. To address this, Quality Sentinel employs CLIP text embeddings of organ classes as conditions to discern the target organ amidst multiple organs within an image. The text embeddings also encourage the model to learn shared representations for similar classes due to their similar conditions. In this way, similar classes can learn similar evaluation criteria and promote each other's performance.

As for the generation of text embeddings, it is crucial to choose an appropriate prompt template \citep{qin2022medical} because it is the key to eliciting knowledge from pre-trained VLMs. According to our ablation study in Section \ref{sect:eval}, the best prompt is ``[CLS]", the class name itself.

Quality Sentinel views text embeddings as class-conditional attention added before the decision layers. For a given class $k$, its CLIP text embedding $\varphi_k$ is concatenated with the image feature $f_1 = E_v(x)$ produced by a vision encoder (\eg ResNet50 \citep{he2016deep}). A multi-layer perceptron (MLP) then computes attention weights $[\omega_1, \omega_2] = MLP([f_1, \varphi_k])$, used by the regression prediction head, consisting of two fully connected (FC) layers $g_1(\cdot)$ and $g_2(\cdot)$, to make the final prediction:
\begin{equation}
\hat{h} = g_2(\omega_2 \otimes g_1(\omega_1 \otimes E_v(x))),
\label{eq2}
\end{equation}
where $\otimes$ denotes element-wise multiplication. Employing two FC layers allows for a phased feature compression, enabling the attention mechanism to more significantly influence prediction outcomes.

\subsection{Optimal Pair Ranking Loss}
Reliance on MSE loss alone is inadequate for training Quality Sentinel due to its tendency to bias DSC predictions towards the most common score interval. Considering the framework's application in ranking sample quality and selecting pseudo labels for revision, we introduce a scale-insensitive ranking loss as an auxiliary training mechanism.

Inspired by the approach in \citep{yoo2019learning}, we compare sample pairs. However, in the context of multi-organ segmentation, comparing dissimilar organ classes (\eg spleen vs. skull) is unproductive. Thus, we form $N/2$ pairs within a batch of size $N$, ensuring each pair consists of identical or similar class samples by maximizing their total embedding similarity. The pairing process involves three steps: 1) computing a cosine similarity matrix $H \in \mathbb{R}^{N\times N}$; 2) converting $H$ into a cost matrix $H' = -H + \infty \cdot I_{N\times N}$ to prevent self-pairing; 3) identifying optimal pairs through the Jonker-Volgenant algorithm for linear sum assignment \citep{crouse2016implementing}.

For an optimal pair $\{x^p=\{x_i, x_j\}\}$, the ranking loss is:
\begin{equation}
    \mathcal{L}_{Rank}(\hat{h}_i,\hat{h}_j,h_i,h_j) = max(0, (\hat{h}_i-\hat{h}_j)\cdot (h_j-h_i) + \xi),
\label{eq3}
\end{equation}
where $\xi$ is a predefined positive margin. This loss specifically penalizes discrepancies in the directional agreement between the predicted and actual DSC differences.

\section{Experiments \& Results}

\subsection{Experimental Settings}
\label{sect: setting}

\textbf{Stage 1 Dataset Construction.} We fine-tuned the pretrained STUNet \citep{huang2023stu} on the DAP Atlas \citep{jaus2023towards}, a comprehensive dataset that aggregates 142 categories from 14 datasets. Model checkpoints were saved at specified epochs: ${10, 20, 30, 40, 50, 100, 200, 300, 400, 500}$. From each checkpoint, pseudo labels for 15 CT scans were generated, with 10 designated for training and 5 for testing, creating a dataset of CT scans paired with pseudo labels of varying quality and their corresponding ground truth DSC. Because DAP Atlas is also an AI-assisted dataset, its ground truth masks may not be accurate enough. Therefore, these imperfect ground truths cannot serve as the training samples with $DSC=1.0$. The generated masks and their DSCs are feasible to train the model because their variations are minimized when averaged over the large dataset. To further decrease the inaccuracy of ground truth DSC for training, we resample data to refine their quality.

\textbf{Stage 2 Data Resampling.} Utilizing the trained Quality Sentinel, we resampled 50 CT scans of the highest label quality from the DAP Atlas to create a new dataset, ensuring a balanced sex distribution (25 male, 25 female) and dividing them into 20 training and 5 testing samples each. This dataset, containing approximately 4 million 2D image-label pairs, has improved training label quality. The stage 2 resampling leads to more accurate ground truth DSC and enhanced model performance, as evidenced in our ablation study in the following section.

\textbf{Training Details.} We opted for ResNet50 as the vision encoder, adjusting its initial layer to accommodate 2-channel inputs. CT HU values were clipped to $[-200,200]$ and images were cropped and resized to $256\times 256$ around pseudo label centers. These processed pseudo-label masks were then concatenated with the CT images for training. The Quality Sentinel was trained over 30 epochs using the Adam optimizer, with a learning rate of $10^{-3}$ and batch size of 128. The loss function incorporated a scaling constant $\lambda=1$. Our experiments were conducted on a PyTorch 2.1.0 framework, utilizing an Intel Xeon Gold 5218R CPU@2.10GHz and 8 Nvidia Quadro RTX 8000 GPUs.

\subsection{Evaluation of Quality Sentinel}
\label{sect:eval}
% 第一个为ablation study表格，第二个为data amount vs performance的图与scatter plot展示性能的图的组合。

\begin{table}[t]
\scriptsize
\centering
\caption{Ablation study for Quality Sentinel. The baseline is a ResNet50 with MSE loss. All results are reported on the resampled testing set.}
\renewcommand{\arraystretch}{1.2} % line height
\resizebox{\textwidth}{!}{%
\begin{tabular}{>{\raggedright\arraybackslash}p{6cm}cc|c|c|c|c}
\toprule
\rule{0pt}{4pt}
Condition~ & ~Opt. Pair Loss~~ & ~Data Resampling~ &  ~~LCC~~ & ~SROCC~ & ~MAP@5~ & ~MAP@10~        \\
\hline
\hline
\rule{0pt}{6pt}
- & - & - & 0.797 & 0.775 & 0.417 & 0.481 \\
\hline
One-hot & - & - & 0.817 & 0.795 & 0.427 & 0.501 \\
CLIP: ``A computerized tomography of a [CLS]." & - & - & 0.840 & 0.805 & 0.438 & 0.510 \\
CLIP: ``There is [CLS] in this computerized tomography."  & - & - & 0.847 & 0.801 & 0.444 & 0.525 \\
CLIP: ``A photo of a [CLS]."  & - & - & 0.852 & 0.808 & 0.445 & 0.533 \\
CLIP: ``[CLS]" & - & - & 0.853 & 0.808 & 0.449 & 0.531 \\
\hline
- & \checkmark & - & 0.817 & 0.778 & 0.438 & 0.535 \\
CLIP: ``[CLS]" & \checkmark & - & 0.855 & 0.810 & 0.463 & 0.546 \\
CLIP: ``[CLS]" & \checkmark & \checkmark & \textbf{0.902} & \textbf{0.856} & \textbf{0.500} & \textbf{0.565} \\
\bottomrule
\end{tabular}
}
\label{tab:ablation}
\end{table}

\textbf{Ablation Study.} To assess the effectiveness of our method, we performed an ablation study concerning text embedding, optimal pair ranking loss, and the data resampling strategy. We utilized three metrics for evaluation: the Linear Correlation Coefficient (LCC), the Spearman Rank Order Correlation Coefficient (SROCC), and the Mean Average Precision (MAP@k), where `k' refers to the samples with the lowest ground truth DSC. 
The MAP@k metric is an important ranking metric in our evaluation as it focuses on the samples with the lowest ground truth Dice Similarity Coefficient (DSC), which are cases that may degrade the model training. For each class, the Average Precision at k (AP@k) is first calculated, which involves selecting the k samples with the lowest actual DSC scores and checking if these indices correspond to low predicted scores as well. The precision is accumulated for hits within the top-k ranked predictions, and the MAP@k is computed as the mean AP@k across all classes.

The ablation study is shown in Tab. \ref{tab:ablation}.  Incorporating each of the proposed techniques significantly enhances model performance.
First, adding class conditions to the model is beneficial for model performance. Even with one-hot embedding of classes, Quality Sentinel performs 2.0\% better than the baseline \wrt LCC. 
For different CLIP embeddings, we have tried four different prompt templates. The results show that the best prompt is the simplest one, ``[CLS]". The text embeddings notably improve both regression (\ie correlation coefficients) and ranking performance (MAP@k). The reason that CLIP embedding performs better than one-hot embedding is that it addresses the label orthogonality problem by exploiting semantic relationships among organs \citep{liu2023clip}. A simpler prompt leads to greater similarity differences between classes, thus eliciting better knowledge from the language.
The optimal pair ranking loss is particularly beneficial for ranking (the MAP@10 improved from 0.481 to 0.535 compared to baseline). This would be helpful when using Quality Sentinel to rank sample quality and filter unsatisfying samples.
Finally, data resampling markedly boosts all evaluated metrics by enriching the dataset both in volume and quality of ground truth labels.

\begin{figure}[t]
\vspace{0cm}                          
\centering\centerline{\includegraphics[width=0.9\linewidth]{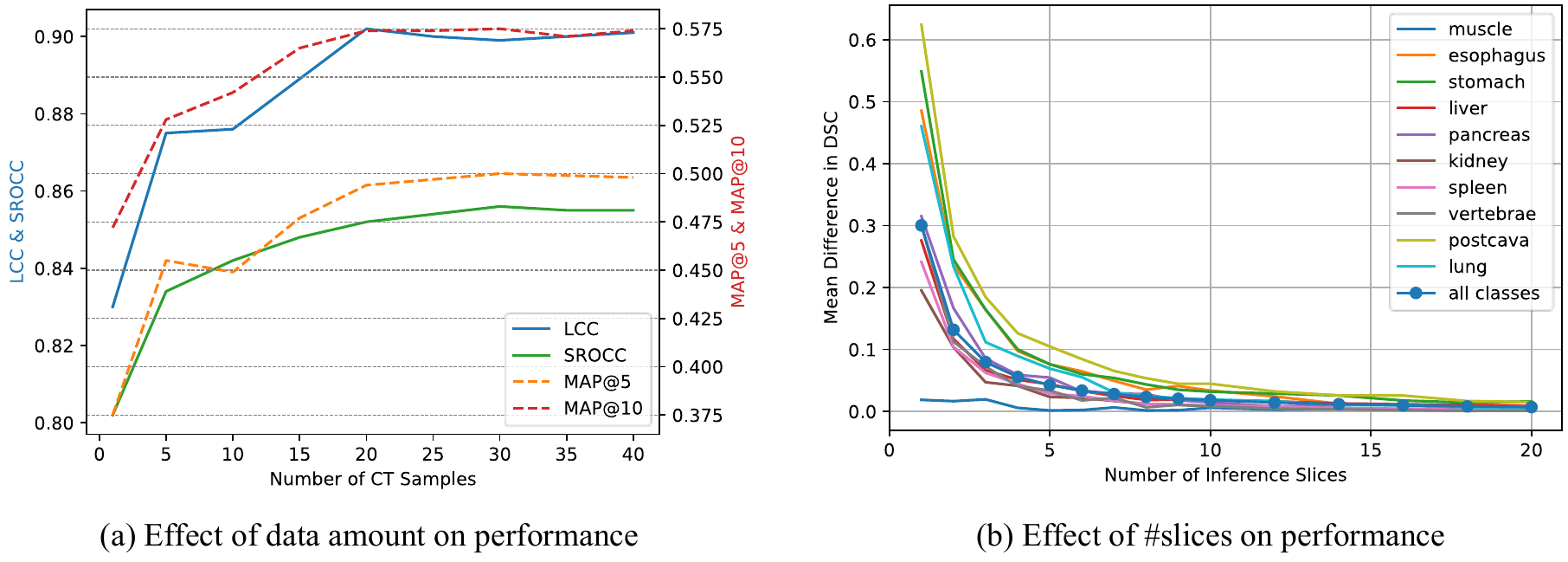}}
\caption{Illustration of (a) the curves of model performance and data amount and (b) the curves of performance deviation \wrt slice amount compared to prediction with all slices.}
\label{fig:effect}
\end{figure}

\textbf{Evaluation of Model Configurations.} The training dataset of Quality Sentinel is obtained from 40 original CT volumes of DAP Atlas. To investigate the impact of training data volume on Quality Sentinel's efficacy, we analyzed how model performance correlates with the amount of data, represented in Fig. \ref{fig:effect} (a). Results indicate performance plateaus at approximately 20-30 CT scans, ensuring the adequacy of our training dataset's size. In addition, the inference of Quality Sentinel uniformly samples 10 2D slices from a 3D mask, we illustrate the mean difference in DSC \wrt number of slices used for inference in Fig. \ref{fig:effect} (b). The difference between sampling strategy and inference with all slices quickly decreases as the number of slices increases. Specifically, when inferring with 10 slices, the mean difference for all classes is only 0.021, which is acceptable for most practical scenarios.

\begin{figure}[t]
\vspace{0cm}                          
\centering\centerline{\includegraphics[width=1.0\linewidth]{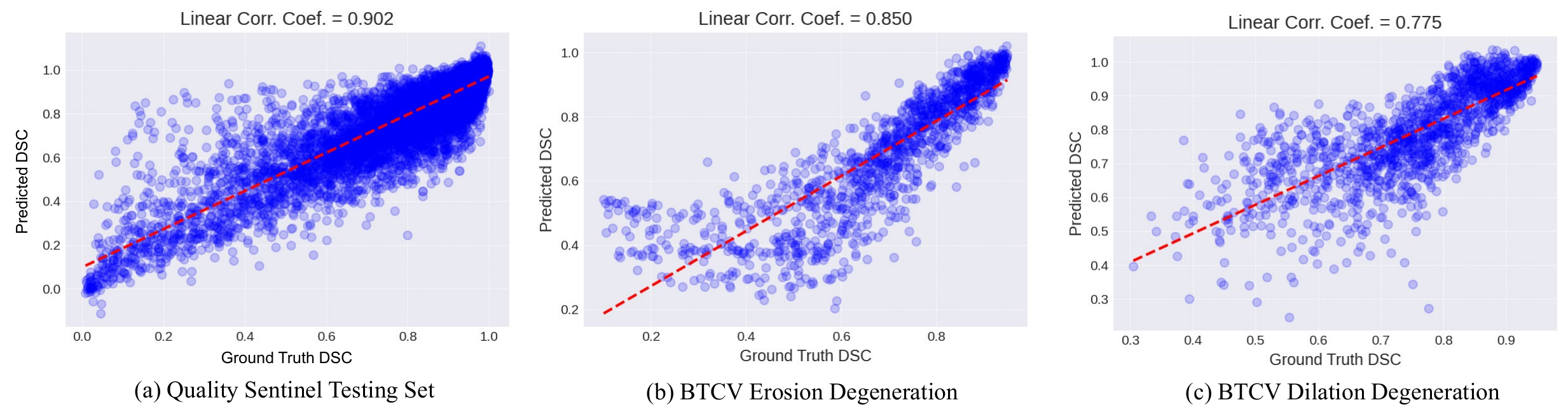}}
\caption{Evaluation of model performance. The scatter plots are illustrated between predicted DSC and ground truth DSC of Quality Sentinel testing set (a) and external evaluation on BTCV with erosion mask degeneration (b) and dilation degeneration (c).}
\label{fig:scatter}
\end{figure}

\begin{figure}[t]
\vspace{0cm}                          
\centering\centerline{\includegraphics[width=1.0\linewidth]{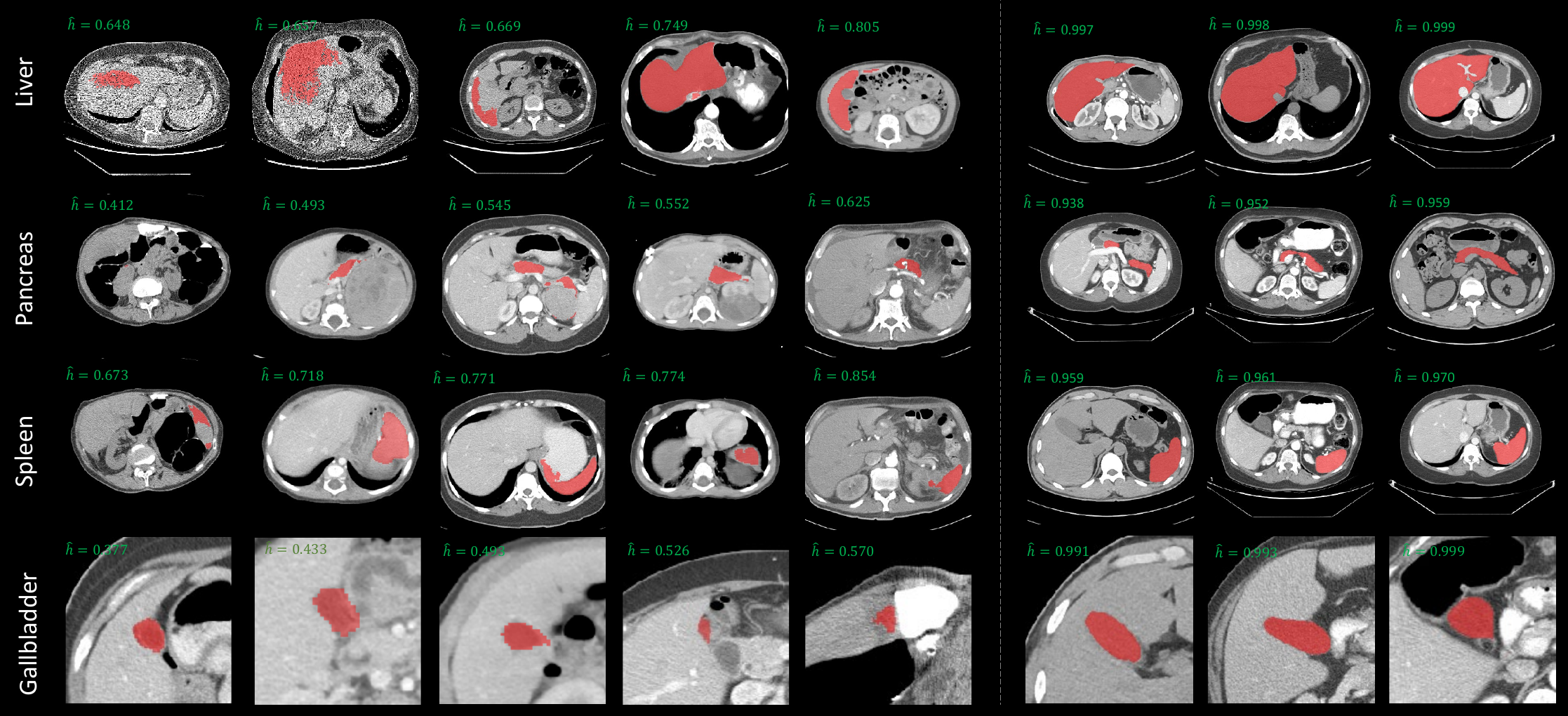}}
\caption{Illustration of labels with lower (left) and higher (right) predicted DSC in AbdomenAtlas dataset. Typical label errors detected include reduced masks (the first two of the liver), missing masks (the first one of the pancreas), and expanded masks (the first three of the gallbladder).}
\label{fig:AAtlas}
\end{figure}

\textbf{External Evaluation.}  Fig. \ref{fig:effect} (a) showcases Quality Sentinel's optimal predictions on its testing set, revealing a strong correlation between the model's DSC predictions and actual values, underscoring the model's predictive reliability. However, generalization risks remain since the model was developed on a single data source. We do external evaluations to validate its performance on more data domains.
First, we use BTCV \citep{landman2015miccai}, a well-known manually annotated dataset with 13 classes, to show the model's performance. As shown in Fig. \ref{fig:effect} (b) and (c), we use morphological operations (\ie erosion and dilation) to generate degraded masks and use Quality Sentinel to infer the DSC, showcasing consistent results with the testing set. Even though the LCC decreases to 0.850 and 0.775 in these two cases, this experiment validates the generalization of Quality Sentinel on out-of-distribution data and unseen mask degradation. We attribute this ability to the large-scale training dataset comprising a wide variety of different degradation modes.
Second, we apply Quality Sentinel to the largest multi-organ segmentation dataset, AbdomenAtlas \citep{qu2024abdomenatlas}, to visually demonstrate the model’s ability to detect label errors. In Fig. \ref{fig:AAtlas}, we show the model prediction of DSCs on four representative abdominal organs. On the left of the figure, imperfect labels with missing, overly expanded, and reduced masks are detected. On the right of the figure, it can be seen that high-quality masks are far smoother and more precise than imperfect masks.

\subsection{Dataset Evaluation with Quality Sentinel}
\label{sect:data_eval}
% DAP Atlas, TotalSegmentator, AbdomenAtlas对比共有的8种器官质量，做表格/箱线图
% 做箱线图展示TotalSegmentator的器官、年龄、性别bias。

\begin{table}[t]
  \centering
  \caption{Comparison of label quality evaluation for different datasets.}
  \label{tab:eval}
  \renewcommand{\arraystretch}{1.2} % line height
  \resizebox{\textwidth}{!}{%
  \begin{tabular}{l|c|c|cccccccc|cc}
    \toprule
    ~~~~~~Dataset & \# Samples & Annotators & \multicolumn{8}{c|}{Organ-wise Average Estimation} & \multicolumn{2}{c}{Overall Estimation}  \\
    \cline{4-13}
      & & & aorta & gallbladder & kidney & liver & pancreas & postcava & spleen & stomach & Mean DSC~ & DSC$<0.8$  \\
    \hline
    \hline
    CHAOS \citep{valindria2018multi} & 20 & Human & - & - & - & \textbf{0.985} & - & - & - & - & \textbf{0.985}  & 0.0\% \\
    TCIA Pancreas \citep{roth2015deeporgan} & 42 & Human & - & 0.904 & 0.963 & 0.961 & 0.919 & - & 0.946 & 0.957 & 0.941  & 0.8\% \\
    BTCV \citep{landman2015miccai} & 47 & Human &  0.943 & 0.875 & 0.961 & 0.981 & 0.928 & 0.921 & 0.962 & 0.969 & 0.944  & 1.7\% \\
    AbdomenCT-12organ \citep{ma2023unleashing} & 50 & Human & \textbf{0.965} & \textbf{0.932} & 0.963 & 0.965 & \textbf{0.961} & \textbf{0.960} & \textbf{0.964} & 0.961 & 0.959 & 0.0\% \\
    WORD \citep{luo2021word} & 120 & Human & - & 0.874 & \textbf{0.968} & 0.981 & 0.946 & - & 0.958 & \textbf{0.970} & 0.952  & 1.9\% \\
    LiTS \citep{bilic2019liver} & 131 & Human & - & - & - & 0.963 & - & - & - & - & 0.963  & 0.0\% \\
    CT-ORG \citep{rister2020ct} & 140 & Human\&AI & - & - & 0.911 & 0.948 & - & - & - & - & 0.924 & 0.5\% \\
    AMOS22 \citep{ji2022amos} & 200 & Human\&AI & 0.954 & 0.874 & 0.956 & 0.971 & 0.928 & 0.927 & 0.953 & 0.942 & 0.941 & 2.1\% \\
    KiTS$^*$ \citep{heller2020international} & 489 & Human & - & - & 0.867 & - & - & - & - & - & 0.867  & 11.3\% \\
    DAP Atlas \citep{jaus2023towards} & 533 & AI & 0.945 & 0.856 & 0.946 & 0.961 & 0.851 & 0.919 & 0.959 & 0.945 & 0.925 & 2.4\% \\
    AbdomenCT-1k \citep{ma2021abdomenct}~ & 1000 & Human\&AI & - & - & 0.951 & 0.963 & 0.949 & - & 0.956 & - & 0.955  & 0.2\% \\
    TotalSegmentator \citep{wasserthal2023totalsegmentator} & 1204 & Human\&AI & 0.927 & 0.794 & 0.914 & 0.941 & 0.853 & 0.880 & 0.933 & 0.915 & 0.901 & 8.0\% \\
    %Abdomen Atlas & 5195 & 0.884 & 0.764 & 0.923 & 0.948 & 0.885 & 0.842 & 0.934 & 0.873 & 0.886 & 12.6\% \\
    %JHH & 5281 & 0.929 & 0.853 & 0.908 & 0.933 & 0.808 & 0.901 & 0.934 & 0.906 & 0.902 & 6.5\% \\
    \bottomrule
  \end{tabular}%
  }
\begin{tablenotes}
\tiny
\item $^*$ Due to a different annotation protocol of the KiTS dataset, its kidney annotation excludes two tumor types included in it and may account for a relatively large volume, resulting in an underestimated DSC.
\end{tablenotes}
\vspace{-0.3cm}
\end{table}

\textbf{Label Quality Comparison.} We conducted a comparative analysis of 12 segmentation datasets focusing on major abdominal organs using Quality Sentinel, as detailed in Tab. \ref{tab:eval}. We observed that smaller datasets, particularly those manually annotated such as AbdomenCT-12organ, a benchmark dataset for the FLARE challenge, exhibit superior organ-wise label quality. Conversely, in larger datasets like TotalSegmentator, where AI-generated pseudo labels are more prevalent, there is a noticeable decline in the overall DSC estimation. Human annotation agreement is usually higher than $0.8$ \citep{qu2024abdomenatlas}. So if considering a quality estimation below $0.8$ as inadequate, Quality Sentinel can be used to diagnose the dataset and find the label errors. For instance, it reveals that the TotalSegmentator contains $8\%$ unsatisfying annotations for the abdomen organs and requires potential refinement.

\begin{figure}[t]
\vspace{0cm}                          
\centering\centerline{\includegraphics[width=0.9\linewidth]{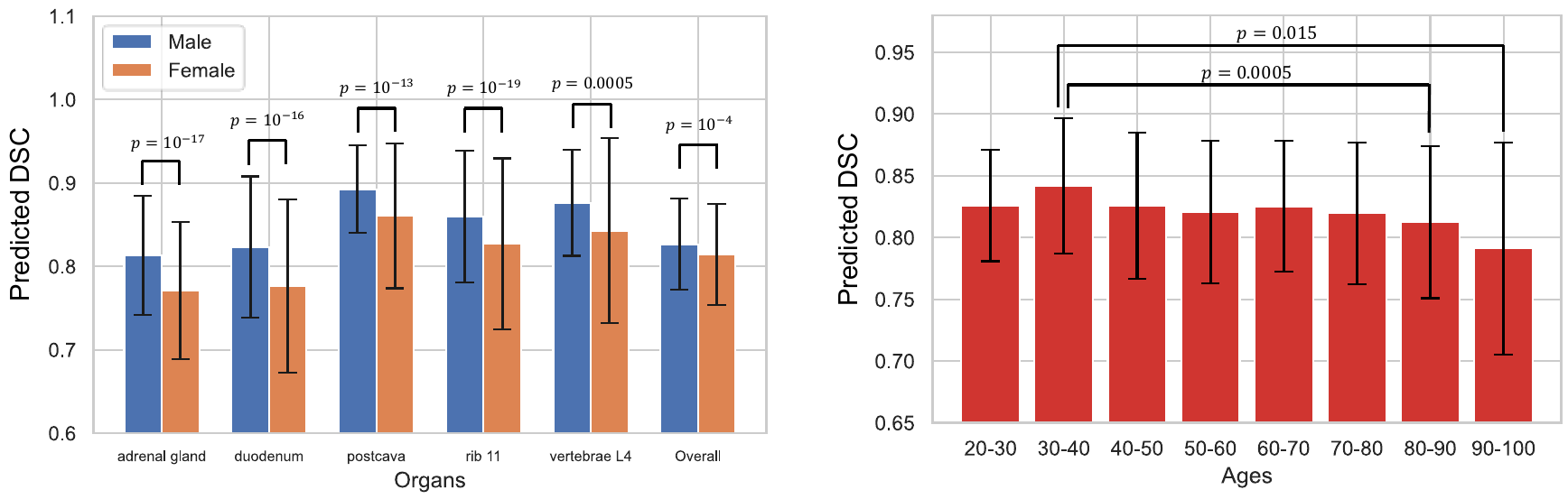}}
\caption{The annotation bias \wrt gender (left) and age (right) for TotalSegmentator. We show several representative organs with higher label quality for males than females and find a significant overall quality difference for different genders. In addition, older patients generally have worse label quality than younger patients.}
\label{fig:bias}
\vspace{-0.3cm}
\end{figure}

\textbf{Annotation Bias.} Investigating the TotalSegmentator dataset, which includes comprehensive metadata, revealed significant biases in label quality related to gender and age, as illustrated in Fig. \ref{fig:bias}. Annotations for males and younger individuals were generally of higher quality compared to females and older individuals, with the discrepancies being statistically significant (gender bias $p<10^{-3}$, age bias $p<0.05$). These biases may stem from the complexity and severity of conditions in certain patients, which can obscure organ visibility and complicate the annotation. To enhance the generalizability of models for clinical application, future dataset constructions should prioritize more meticulous annotations for these groups.

\subsection{Data Efficient Training}
\label{sect:efficient_training}
% 第一个为self-training表格，第二个为active learning的折线图。

In this section, we demonstrate that Quality Sentinel effectively ranks/selects training samples, enhancing the efficiency of training segmentation models. We assessed its performance against random selection, Monte-Carlo Dropout (for diversity) \citep{gal2016dropout}, and Entropy (for uncertainty) \citep{joshi2009multi} on the TotalSegmentator dataset \citep{wasserthal2023totalsegmentator}. We chose this dataset for its wide range of 104 classes and it is a fair out-of-domain benchmark for comparison. Utilizing a pretrained SwinUNETR \citep{tang2022self}, we fine-tuned the model on 20 full-body CT scans and generated pseudo labels for an additional 100 samples. We then employed these methods to evaluate the quality of pseudo labels, selecting the most suitable samples for further annotating/training, with DSC and Normalized Surface Distance (NSD) serving as the key performance indicators. In the following experiments, the only difference between groups is their selected samples based on different methods.

\begin{figure}[t]
\vspace{0cm}                          
\centering\centerline{\includegraphics[width=0.9\linewidth]{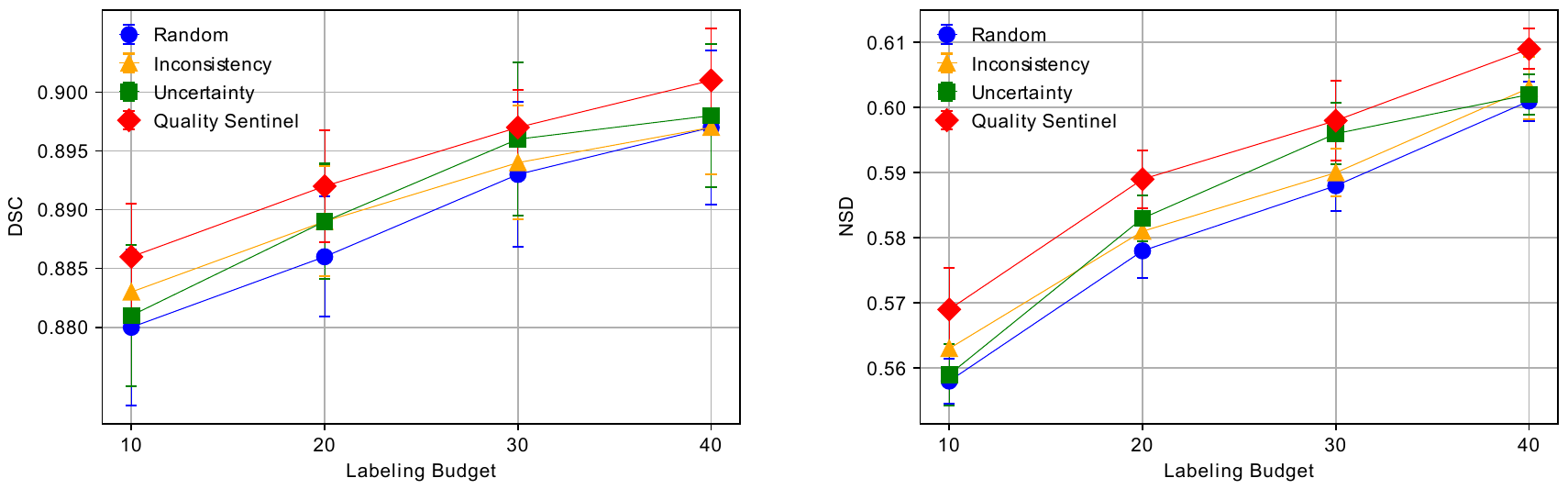}}
\caption{Human-in-the-Loop (active learning) results of label quality ranking methods on the TotalSegmentator. Error bars are reported with 5 repeated experiments. Quality Sentinel helps to reduce annotation costs from 30 samples to 20 samples compared to random selection.}
\label{fig:active}
\vspace{-0cm} 
\end{figure}

\textbf{Human-in-the-Loop (Active Learning).} In the human-in-the-loop annotation of datasets, we usually have a small annotation budget, and select a small number of samples that the initial model cannot predict well. These identified samples (namely most informative samples in active learning) are then annotated with human annotators. In our experiment, each method selects $n$ samples based on the lowest pseudo-label quality and revises them with ground truth labels. We combine these samples alongside the original 20 samples for fine-tuning segmentation models. Results (Fig. \ref{fig:active}) reveal that Quality Sentinel consistently outperforms the other methods. With a labeling budget of 20 samples, it exceeds random selection by 0.006 and 0.011 in DSC and NSD metrics, respectively. This is comparable to a random selection budget of 30 samples. It means when used for real annotation scenarios, Quality Sentinel can help to reduce about 1/3 annotation cost.

\textbf{Semi-supervised Learning.} In this experiment, we perform a typical semi-supervised learning strategy: self-training with pseudo labels. Each ranking method selects the top 20 pseudo labels with the highest quality. We combine these samples and the initial 20 ground truth-labeled samples to fine-tune The segmentation models. If the selection is not reliable enough and low-quality pseudo-labels are included for training, the model performance may not improve or even be harmed. As Tab.~\ref{tab:semi} indicates, Quality Sentinel outperforms all alternatives. Specifically, it increases the performance improvement by 33\% ($0.67\% \rightarrow 0.89\%$) on DSC and 88\% ($0.41\% \rightarrow 0.77\%$) on NSD compared to the entropy-based method. Compared to random selection, Quality Sentinel has statistically significant improvements ($p=0.036$ for DSC and $p=0.0035$ for NSD). 
As for resource overhead, MC dropout and entropy-based methods need to operate on the 3D raw outputs of the model with the shape $[C,D,H,W]$ and \textit{float} data type. Entropy needs to be computed with per-voxel logits, and MC dropout requires the segmentation model to do inference three times and calculate the standard deviation of each voxel's probability. In practice, they both require saving the model's 3D outputs to disk and reloading them to memory for further computation. Instead, Quality Sentinel only needs the segmentation masks with the shape $[D,H,W]$ and the data type of \textit{uint8}. Considering their different compression formats (.npy and .nii.gz), the disk usage can vary by 20,000 times. The 2D processing style and the uniform sampling scheme of Quality Sentinel also largely reduce quality estimation costs (6 times less time and 60 times less RAM than MC dropout).

\begin{table}[t]
  \centering
  \caption{Semi-supervised learning results of label quality ranking methods on the TotalSegmentator dataset. Standard deviations are reported with 5 repeated experiments.}
  \label{tab:semi}
  \resizebox{\textwidth}{!}{%
  \begin{tabular}{l|cc|cc|ccc}
    \toprule
    ~~~~Method & \multicolumn{2}{c|}{\# Scans Used} & \multicolumn{2}{c|}{Metrics} & \multicolumn{3}{c}{Quality Estimation Cost} \\
    \cline{2-8}
     & ~Labeled~ & ~Unlabeled~ & DSC[\%] & NSD[\%] & ~Time[h]~ & ~RAM[GB]~ & ~Disk[GB]~ \\
    \hline
    SwinUNETR~~ & 20 & 0 & 82.55 ($\pm$ 0.31) & 45.76 ($\pm$ 0.28) & - & - & - \\
    \hline
    \hline
    Random & 20 & 100 & 83.16 ($\pm$ 0.26) & 46.08 ($\pm$ 0.33) & - & - & - \\
    MC dropout & 20 & 100 & 83.32 ($\pm$ 0.57) & 46.35 ($\pm$ 0.46) & 3.17 & 92.4 & 1032 \\
    Entropy & 20 & 100 & 83.21 ($\pm$ 0.37) & 46.17 ($\pm$ 0.43) & 1.61 & 33.1 & 344 \\
    Quality Sentinel & 20 & 100 & \textbf{83.44} ($\pm$ 0.35) & \textbf{46.53} ($\pm$ 0.38) & 0.50 & 1.5 & 0.05 \\
    \bottomrule
  \end{tabular}%
  }
\end{table}

\section{Discussion and Limitations}
\label{sect: limitations}
Quality Sentinel learns generalizable evaluation capability on a large dataset comprising 4M slices. Although it shows good performance on external evaluation, it may not cover all segmentation scenarios. Its training data come from the DAP Atlas dataset with 142 healthy organs. In the KiTS dataset, kidney and kidney tumors are marked as two categories. The definition discrepancy makes the model underestimate label quality (Tab. \ref{tab:eval}). Moreover, Quality Sentinel is trained on a dataset with balanced gender. Due to the annotation bias discussed before, the different annotation qualities of different groups would possibly affect the ground truth DSC during training. Finally, with Quality Sentinel, it is possible to achieve purposeful QC in data sets. On the one hand, it can help improve the annotation and balance of the data set, but on the other hand, there is also the risk of being exploited to intentionally control the preferences of the data and model.

\section{Conclusion}

Our primary contribution is the development of Quality Sentinel, a label quality estimation model trained on a vast and diverse set of image-label pairs. This model enables a thorough evaluation of dataset quality and identification of biases. It acts as a pivotal tool for sample selection, both for guiding human-in-the-loop annotation and for model training. While our empirical results show the practical value of the model on CT images, an important future work is to extend Quality Sentinel to more modalities to be a generic label evaluator for medical image segmentation. We envision that Quality Sentinel will play a crucial role in quality control for expansive datasets, thereby enhancing training efficiency and applicability in real-world scenarios.

\clearpage
\bibliographystyle{plainnat}
\bibliography{CCQN}

\end{document}